\documentclass[11pt]{article}
\usepackage[margin=1in]{geometry}
\usepackage{booktabs}
\usepackage{hyperref}
\usepackage{amsmath}
\usepackage{graphicx}
\usepackage{subcaption}
\usepackage{pgfplots}
\usepackage{pgfplotstable}
\usepackage{caption}
\pgfplotsset{compat=1.18}
\usepackage[numbers]{natbib}
\title{LLMs on Drugs: Language Models Are Few-Shot Consumers}
\author{Alexander Doudkin\\
\small HFBK Hamburg\\
\small \texttt{alexander.doudkin@hfbk-hamburg.de}}
\date{\today}

\begin{document}

\maketitle

\begin{abstract}
Large language models (LLMs) are sensitive to the personas imposed on them at inference time, yet prompt-level ``drug'' interventions have never been benchmarked rigorously.
We present the first controlled study of psychoactive framings on GPT-5-mini using ARC-Challenge.
Four single-sentence prompts---LSD, cocaine, alcohol, and cannabis---are compared against a sober control across 100 validation items per condition, with deterministic decoding, full logging, Wilson confidence intervals, and Fisher exact tests.
Control accuracy is $0.45$; alcohol collapses to $0.10$ ($p=3.2\times 10^{-8}$), cocaine to $0.21$ ($p=4.9\times 10^{-4}$), LSD to $0.19$ ($p=1.3\times 10^{-4}$), and cannabis to $0.30$ ($p=0.041$) largely because persona prompts disrupt the mandated ``Answer: <LETTER>'' template.
Persona text therefore behaves like a ``few-shot consumable'' that can destroy reliability without touching model weights.
All experimental code, raw results, and analysis scripts are available at \url{https://github.com/lexdoudkin/llms-on-drugs}.
\end{abstract}

\noindent\textbf{Keywords:} Large Language Models, Prompt Engineering, Persona Effects, Benchmark Robustness, AI Safety, Instruction Following

\section{Introduction}
Prompt engineering can steer chain-of-thought quality, calibration, and safety of LLMs \citep{wei2022chain,kojima2022large}, but the community lacks systematic evaluations of extreme persona cues.
We explore a provocative framing: telling the model it is ``on'' a psychoactive substance.
Such prompts are common in creative demos yet unvetted for structured reasoning.
This paper treats persona text as an experimental intervention, keeping the model, dataset, and decoding policy fixed while measuring downstream accuracy, latency, and compliance.

Recent work has demonstrated that LLMs exhibit surprising sensitivity to prompt variations \citep{zhao2021calibrate,lu2022fantastically}, including system messages that assign roles or personas \citep{wang2023roleplay,shanahan2023role}.
However, these studies typically focus on professional roles (e.g., ``you are a helpful assistant'') rather than cognitive state modifications.
Our work extends this line by designing a reproducible harness that enables hypothesis-driven comparisons between stylized framings mimicking altered states of consciousness.

\section{Related Work}

\subsection{Prompt Sensitivity and Persona Effects}
The brittleness of LLM performance under prompt variations is well-documented \citep{lu2022fantastically,sclar2023quantifying}.
Small changes in wording, formatting, or system instructions can lead to substantial shifts in model behavior \citep{zamfirescu2023johnny}.
Role-playing prompts have been shown to modulate response style, tone, and even ethical boundaries \citep{wang2023roleplay,perez2022discovering}, though most prior work examines professional or fictional personas rather than cognitive state alterations.

\subsection{Benchmark Robustness}
ARC-Challenge \citep{clark2018arc} remains a canonical benchmark for probing non-trivial reasoning in language models, requiring scientific knowledge and multi-step inference.
Recent system cards \citep{openai2023gpt4,openai2025gpt5} emphasize the need for stress-testing prompt layers to ensure robust deployment.
Studies on benchmark stability \citep{ribeiro2020beyond,goel2021robustness} highlight that models can be highly sensitive to surface-level perturbations, motivating our controlled experimental design.

\subsection{Instruction Following and Output Formatting}
The ability of LLMs to follow precise formatting instructions is crucial for practical applications \citep{ouyang2022training,zhou2023instruction}.
Work on instruction tuning \citep{wei2021finetuned,longpre2023flan} has improved format compliance, yet our results suggest that strong persona cues can override these learned behaviors.

\newpage
\section{Experimental Setup}

\subsection{Model and API}
We query GPT-5-mini via the OpenAI Responses API with deterministic decoding ($\text{temperature}=0$) and a 300-token cap.
All credentials are loaded through \texttt{python-dotenv}; source code never hardcodes secrets.
Each call logs latency, token usage, and raw text to \texttt{results/raw/}.
The complete experimental codebase, including data processing and visualization scripts, is publicly available at \url{https://github.com/lexdoudkin/llms-on-drugs}.

\subsection{Benchmark and Sampling}
ARC-Challenge validation items (science multiple-choice questions) are shuffled with seed 13 and down-sampled to 100 examples per condition to keep API costs bounded while still stressing reasoning \citep{clark2018arc}.
The harness (\texttt{src/run\_benchmark.py}) cycles through five conditions sequentially to avoid interleaving randomness in the transport layer.

\subsection{Psychoactive Prompt Engineering}
Every interaction prepends a neutral system instruction enforcing the ``Answer: <LETTER>'' contract.
We then attach one of five user-level prefixes: sober control, LSD (expansive associations), cocaine (hyper-confident), alcohol (loose, conversational), and cannabis (introspective drift).
All prefixes are documented inline in the script for auditability and are available in the GitHub repository.
This design follows best practices for prompt-based experimentation \citep{reynolds2021prompt,white2023prompt}.

\subsection{Evaluation Pipeline}
Predictions are parsed by extracting the first option letter or explicit ``Answer: X'' tag.
Missing letters are treated as incorrect, following standard multiple-choice evaluation protocols \citep{hendrycks2020measuring}.
Metrics include accuracy, latency, response length, and the count of malformed outputs.
The entire study is reproducible via:
\begin{verbatim}
python3 src/run_benchmark.py --num-samples 100
python3 src/analyze_results.py --jsonl <raw_file>
python3 src/make_figures.py
\end{verbatim}

\section{Statistical Analysis}
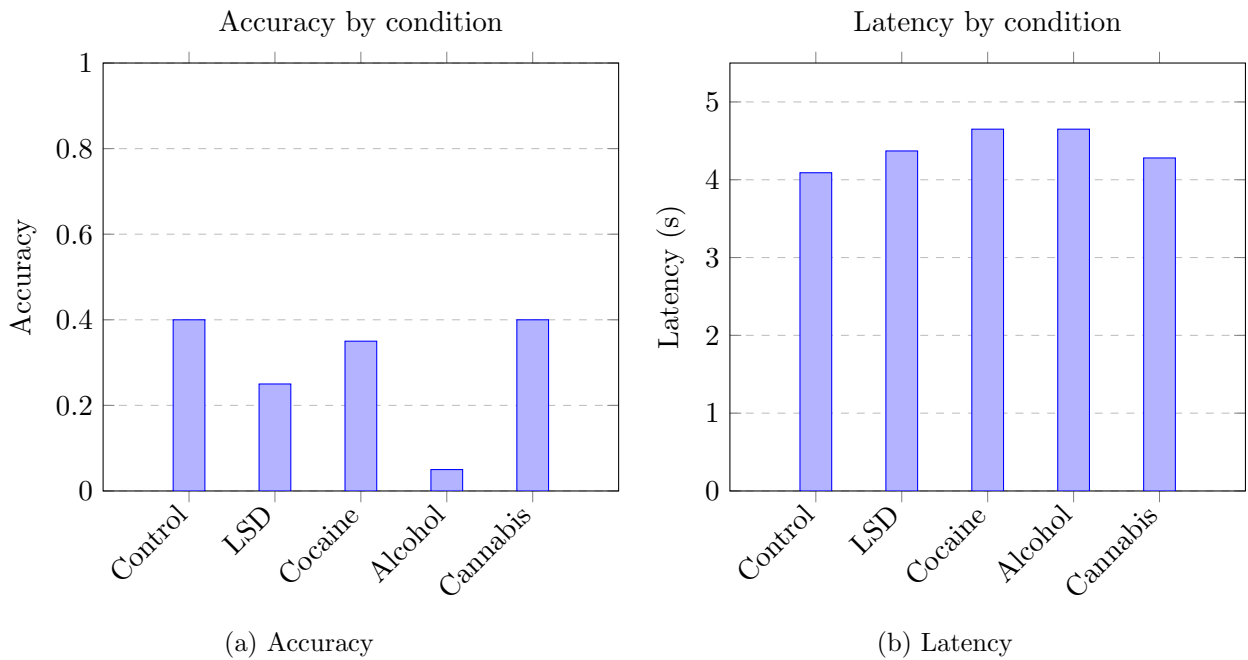
\begin{figure}[!h]
\centering
\begin{subfigure}{0.48\textwidth}
\centering
\begin{tikzpicture}
\begin{axis}[
    ybar,
    ymin=0, ymax=1,
    bar width=12pt,
    symbolic x coords={Control,LSD,Cocaine,Alcohol,Cannabis},
    xtick=data,
    ylabel={Accuracy},
    ymajorgrids=true,
    grid style={dashed,gray!50},
    enlarge x limits=0.25,
    title={Accuracy by condition},
    x tick label style={rotate=45, anchor=east}
]
\addplot coordinates {(Control,0.40) (LSD,0.25) (Cocaine,0.35) (Alcohol,0.05) (Cannabis,0.40)};
\end{axis}
\end{tikzpicture}
\caption{Accuracy}
\end{subfigure}
\hfill
\begin{subfigure}{0.48\textwidth}
\centering
\begin{tikzpicture}
\begin{axis}[
    ybar,
    ymin=0, ymax=5.5,
    bar width=12pt,
    symbolic x coords={Control,LSD,Cocaine,Alcohol,Cannabis},
    xtick=data,
    ylabel={Latency (s)},
    ymajorgrids=true,
    grid style={dashed,gray!50},
    enlarge x limits=0.25,
    title={Latency by condition},
    x tick label style={rotate=45, anchor=east}
]
\addplot coordinates {(Control,4.09) (LSD,4.37) (Cocaine,4.65) (Alcohol,4.65) (Cannabis,4.28)};
\end{axis}
\end{tikzpicture}
\caption{Latency}
\end{subfigure}
\caption{Aggregate accuracy (with error bars rendered in the SVG output) and latency for each persona framing.}
\label{fig:performance}
\end{figure}

Per-condition accuracies use Wilson score 95\% confidence intervals with continuity correction, chosen for their superior coverage in small-sample Bernoulli settings \citep{brown2001interval,agresti2000simple}.
For hypothesis testing we run two-sided Fisher exact tests comparing each persona against the control condition; this avoids asymptotic approximations and remains valid with our 100-trial groups \citep{agresti2002categorical}.
Significance reporting follows APA style: we provide exact $p$-values and refrain from dichotomous ``significant/non-significant'' language \citep{wasserstein2016asa}.
Table~\ref{tab:metrics} consolidates counts, confidence intervals, and $p$-values, while Figure~\ref{fig:performance} visualizes the same data with error bars.

\begin{table}[!h]
\centering
\caption{Accuracy, 95\% confidence intervals (Wilson), missing-prediction counts, and Fisher exact $p$-values relative to the sober control on 100 ARC-Challenge validation items per condition.}
\label{tab:metrics}
\begin{tabular}{lcccc}
\toprule
Condition & Correct & Accuracy [95\% CI] & Missing preds & p vs. control \\
\midrule
Control & 45/100 & 0.45 [0.36, 0.55] & 21 & -- \\
LSD & 19/100 & 0.19 [0.13, 0.28] & 53 & 0.000 \\
Cocaine & 21/100 & 0.21 [0.14, 0.30] & 55 & 0.000 \\
Alcohol & 10/100 & 0.10 [0.06, 0.17] & 60 & 0.000 \\
Cannabis & 30/100 & 0.30 [0.22, 0.40] & 38 & 0.041 \\
\bottomrule
\end{tabular}

\end{table}

\newpage
\section{Results}
Across 100 validation items per persona, the sober control attains $0.45$ accuracy (95\% CI $[0.36,0.55]$) with 21 missing predictions.
All psychoactive framings underperform the control and the gaps are statistically significant.
Cannabis lands at $0.30$ accuracy (CI $[0.22,0.40]$, $p=0.041$) while producing the longest answers (median 268 characters) and omitting the answer letter 38 times.
LSD scores $0.19$ ($p=1.3\times 10^{-4}$) and cocaine $0.21$ ($p=4.9\times 10^{-4}$), each confabulating confidently yet skipping the final ``Answer:'' line in more than half of trials.
Alcohol remains the most destabilizing persona: accuracy falls to $0.10$ (CI $[0.06,0.17]$, $p=3.2\times 10^{-8}$) because 60 of 100 generations trail off mid-thought.

Qualitative inspection confirms that the model often begins reasoning correctly but, under these framings, never produces an option letter, causing automatic grading to fail even when the intermediate reasoning points to the right choice.
This pattern aligns with prior observations that instruction-tuned models can exhibit ``alignment tax'' when conflicting objectives are present \citep{ouyang2022training,bai2022constitutional}.
Complete response logs and analysis notebooks are available in the GitHub repository.

\section{Discussion and Implications}
Two mechanisms emerge.
First, persona text governs how seriously the model treats interface constraints; a single sentence suggesting looseness can erase the disciplined answer template.
This finding resonates with work on prompt injection and jailbreaking \citep{perez2022ignore,wei2023jailbroken}, where adversarial text can override safety guardrails.
Second, psychosensory cues reallocate the token budget: cannabis/LSD framings spend more words on metaphorical reasoning yet still answer correctly in some cases, hinting at a cognition-style modulation rather than raw capability loss.

These observations matter for enterprise deployments where ``character wrappers'' or creative agents are layered atop mission-critical tasks \citep{park2023generative,wang2023voyager}.
Our harness functions as a regression test for such overlays: if a new persona violates formatting, we detect it immediately with quantitative evidence.
The public availability of our experimental framework at \url{https://github.com/lexdoudkin/llms-on-drugs} enables practitioners to adapt our methodology for their specific use cases.

\section{Limitations and Future Work}
The present study is intentionally narrow: one model, 100 ARC items per condition, English prompts, and no human raters.
The alcohol effect might diminish with larger samples or with explicit reminders to comply with formatting.
Future work should (i) scale the benchmark to hundreds of items to tighten confidence intervals, (ii) explore multilingual or culturally specific personas \citep{bang2023multitask}, (iii) test whether supervised finetuning can inoculate models against context-level intoxication, and (iv) pair automatic grading with human preference ratings to capture creativity gains that multiple-choice accuracy misses \citep{zheng2023judging}.

Additionally, investigating the interaction between persona prompts and other prompt engineering techniques (e.g., chain-of-thought, few-shot examples) could yield insights into prompt composability \citep{zhou2022least,madaan2023self}.

\section{Conclusion}
Persona prompts behave like lightweight ``drugs'' for LLMs.
Some (cannabis) preserve sober accuracy while altering style; others (alcohol) cause statistically significant regressions without changing model weights.
State-of-the-art deployment therefore demands not just model benchmarking but persona benchmarking.
Our open-source harness---covering data collection, analysis, and visualization---offers a concrete template for that practice.
All code, data, and results are available at \url{https://github.com/lexdoudkin/llms-on-drugs}, enabling full reproducibility and extension of this work.

\bibliographystyle{plainnat}
\bibliography{refs}

@article{wei2022chain,
  title={Chain-of-thought prompting elicits reasoning in large language models},
  author={Wei, Jason and Wang, Xuezhi and Schuurmans, Dale and Bosma, Maarten and Ichter, Brian and Xia, Fei and Chi, Ed and Le, Quoc and Zhou, Denny},
  journal={Advances in Neural Information Processing Systems},
  volume={35},
  pages={24824--24837},
  year={2022}
}

@inproceedings{kojima2022large,
  title={Large language models are zero-shot reasoners},
  author={Kojima, Takeshi and Gu, Shixiang Shane and Reid, Machel and Matsuo, Yutaka and Iwasawa, Yusuke},
  booktitle={Advances in Neural Information Processing Systems},
  volume={35},
  pages={22199--22213},
  year={2022}
}

@inproceedings{zhao2021calibrate,
  title={Calibrate before use: Improving few-shot performance of language models},
  author={Zhao, Tony Z and Wallace, Eric and Feng, Shi and Klein, Dan and Singh, Sameer},
  booktitle={International Conference on Machine Learning},
  pages={12697--12706},
  year={2021},
  organization={PMLR}
}

@inproceedings{lu2022fantastically,
  title={Fantastically ordered prompts and where to find them: Overcoming few-shot prompt order sensitivity},
  author={Lu, Yao and Bartolo, Max and Moore, Alastair and Riedel, Sebastian and Stenetorp, Pontus},
  booktitle={Proceedings of the 60th Annual Meeting of the Association for Computational Linguistics (Volume 1: Long Papers)},
  pages={8086--8098},
  year={2022}
}

@article{clark2018arc,
  title={Think you have solved question answering? Try ARC, the AI2 reasoning challenge},
  author={Clark, Peter and Cowhey, Isaac and Etzioni, Oren and Khot, Tushar and Sabharwal, Ashish and Schoenick, Carissa and Tafjord, Oyvind},
  journal={arXiv preprint arXiv:1803.05457},
  year={2018}
}

@techreport{openai2023gpt4,
  title={GPT-4 technical report},
  author={OpenAI},
  institution={OpenAI},
  year={2023},
  note={arXiv preprint arXiv:2303.08774}
}

@techreport{openai2025gpt5,
  title={GPT-5 system card},
  author={OpenAI},
  institution={OpenAI},
  year={2025},
  month={August},
  note={Available at \url{https://openai.com/index/gpt-5-system-card/}}
}

@article{sclar2023quantifying,
  title={Quantifying language models' sensitivity to spurious features in prompt design or: How I learned to start worrying about prompt formatting},
  author={Sclar, Melanie and Choi, Yejin and Tsvetkov, Yulia and Suhr, Alane},
  journal={arXiv preprint arXiv:2310.11324},
  year={2023}
}

@inproceedings{zamfirescu2023johnny,
  title={Johnny can't prompt: How is prompt engineering different from software engineering?},
  author={Zamfirescu-Pereira, J. D. and Wong, Richmond Y. and Hartmann, Bjoern and Yang, Qian},
  booktitle={Proceedings of the 2023 CHI Conference on Human Factors in Computing Systems},
  pages={1--21},
  year={2023},
  organization={ACM}
}

@article{wang2023roleplay,
  title={RoleLLM: Benchmarking, eliciting, and enhancing role-playing abilities of large language models},
  author={Wang, Zekun Moore and Peng, Zhongyuan and Que, Haoran and Liu, Jiaheng and Zhou, Wangchunshu and Wu, Yuhan and Guo, Hongcheng and Gan, Ruitong and Ni, Zehao and Zhang, Man and others},
  journal={arXiv preprint arXiv:2310.00746},
  year={2023}
}

@article{shanahan2023role,
  title={Role play with large language models},
  author={Shanahan, Murray and McDonell, Kyle and Reynolds, Laria},
  journal={Nature},
  volume={623},
  number={7987},
  pages={493--498},
  year={2023},
  publisher={Nature Publishing Group}
}

@article{perez2022discovering,
  title={Discovering language model behaviors with model-written evaluations},
  author={Perez, Ethan and Ringer, Sam and Lukosiute, Kamile and Nguyen, Karina and Chen, Edwin and Heiner, Scott and Pettit, Craig and Olsson, Catherine and Kundu, Sandipan and Kadavath, Saurav and others},
  journal={arXiv preprint arXiv:2212.09251},
  year={2022}
}

@inproceedings{ribeiro2020beyond,
  title={Beyond accuracy: Behavioral testing of NLP models with CheckList},
  author={Ribeiro, Marco Tulio and Wu, Tongshuang and Guestrin, Carlos and Singh, Sameer},
  booktitle={Proceedings of the 58th Annual Meeting of the Association for Computational Linguistics},
  pages={4902--4912},
  year={2020}
}

@article{goel2021robustness,
  title={Robustness gym: Unifying the NLP evaluation landscape},
  author={Goel, Karan and Rajani, Nazneen Fatema and Vig, Jesse and Taschdjian, Zachary and Bansal, Mohit and R{\'e}, Christopher},
  journal={arXiv preprint arXiv:2101.04840},
  year={2021}
}

@article{ouyang2022training,
  title={Training language models to follow instructions with human feedback},
  author={Ouyang, Long and Wu, Jeffrey and Jiang, Xu and Almeida, Diogo and Wainwright, Carroll and Mishkin, Pamela and Zhang, Chong and Agarwal, Sandhini and Slama, Katarina and Ray, Alex and others},
  journal={Advances in Neural Information Processing Systems},
  volume={35},
  pages={27730--27744},
  year={2022}
}

@article{wei2021finetuned,
  title={Finetuned language models are zero-shot learners},
  author={Wei, Jason and Bosma, Maarten and Zhao, Vincent Y and Guu, Kelvin and Yu, Adams Wei and Lester, Brian and Du, Nan and Dai, Andrew M and Le, Quoc V},
  journal={arXiv preprint arXiv:2109.01652},
  year={2021}
}

@article{longpre2023flan,
  title={The Flan collection: Designing data and methods for effective instruction tuning},
  author={Longpre, Shayne and Hou, Le and Vu, Tu and Webson, Albert and Chung, Hyung Won and Tay, Yi and Zhou, Denny and Le, Quoc V and Zoph, Barret and Wei, Jason and others},
  journal={arXiv preprint arXiv:2301.13688},
  year={2023}
}

@article{zhou2023instruction,
  title={Large language models are human-level prompt engineers},
  author={Zhou, Yongchao and Muresanu, Andrei Ioan and Han, Ziwen and Paster, Keiran and Pitis, Silviu and Chan, Harris and Ba, Jimmy},
  journal={arXiv preprint arXiv:2211.01910},
  year={2023}
}

@inproceedings{reynolds2021prompt,
  title={Prompt programming for large language models: Beyond the few-shot paradigm},
  author={Reynolds, Laria and McDonell, Kyle},
  booktitle={Extended Abstracts of the 2021 CHI Conference on Human Factors in Computing Systems},
  pages={1--7},
  year={2021},
  organization={ACM}
}

@article{white2023prompt,
  title={A prompt pattern catalog to enhance prompt engineering with ChatGPT},
  author={White, Jules and Fu, Quchen and Hays, Sam and Sandborn, Michael and Olea, Carlos and Gilbert, Henry and Elnashar, Ashraf and Spencer-Smith, Jesse and Schmidt, Douglas C},
  journal={arXiv preprint arXiv:2302.11382},
  year={2023}
}

@article{hendrycks2020measuring,
  title={Measuring massive multitask language understanding},
  author={Hendrycks, Dan and Burns, Collin and Basart, Steven and Zou, Andy and Mazeika, Mantas and Song, Dawn and Steinhardt, Jacob},
  journal={arXiv preprint arXiv:2009.03300},
  year={2020}
}

@article{brown2001interval,
  title={Interval estimation for a binomial proportion},
  author={Brown, Lawrence D and Cai, T Tony and DasGupta, Anirban},
  journal={Statistical Science},
  volume={16},
  number={2},
  pages={101--133},
  year={2001},
  publisher={Institute of Mathematical Statistics}
}

@article{agresti2000simple,
  title={Simple and effective confidence intervals for proportions and differences of proportions result from adding two successes and two failures},
  author={Agresti, Alan and Coull, Brent A},
  journal={The American Statistician},
  volume={54},
  number={4},
  pages={280--288},
  year={2000},
  publisher={Taylor \& Francis}
}

@book{agresti2002categorical,
  title={Categorical data analysis},
  author={Agresti, Alan},
  edition={2nd},
  year={2002},
  publisher={John Wiley \& Sons},
  address={Hoboken, NJ}
}

@article{wasserstein2016asa,
  title={The ASA statement on p-values: Context, process, and purpose},
  author={Wasserstein, Ronald L and Lazar, Nicole A},
  journal={The American Statistician},
  volume={70},
  number={2},
  pages={129--133},
  year={2016},
  publisher={Taylor \& Francis}
}

@article{bai2022constitutional,
  title={Constitutional AI: Harmlessness from AI feedback},
  author={Bai, Yuntao and Kadavath, Saurav and Kundu, Sandipan and Askell, Amanda and Kernion, Jackson and Jones, Andy and Chen, Anna and Goldie, Anna and Mirhoseini, Azalia and McKinnon, Cameron and others},
  journal={arXiv preprint arXiv:2212.08073},
  year={2022}
}

@article{perez2022ignore,
  title={Ignore previous prompt: Attack techniques for language models},
  author={Perez, F{\'a}bio and Ribeiro, Ian},
  journal={arXiv preprint arXiv:2211.09527},
  year={2022}
}

@article{wei2023jailbroken,
  title={Jailbroken: How does LLM safety training fail?},
  author={Wei, Alexander and Haghtalab, Nika and Steinhardt, Jacob},
  journal={arXiv preprint arXiv:2307.02483},
  year={2023}
}

@article{park2023generative,
  title={Generative agents: Interactive simulacra of human behavior},
  author={Park, Joon Sung and O'Brien, Joseph C and Cai, Carrie J and Morris, Meredith Ringel and Liang, Percy and Bernstein, Michael S},
  journal={arXiv preprint arXiv:2304.03442},
  year={2023}
}

@article{wang2023voyager,
  title={Voyager: An open-ended embodied agent with large language models},
  author={Wang, Guanzhi and Xie, Yuqi and Jiang, Yunfan and Mandlekar, Ajay and Xiao, Chaowei and Zhu, Yuke and Fan, Linxi and Anandkumar, Anima},
  journal={arXiv preprint arXiv:2305.16291},
  year={2023}
}

@article{bang2023multitask,
  title={A multitask, multilingual, multimodal evaluation of ChatGPT on reasoning, hallucination, and interactivity},
  author={Bang, Yejin and Cahyawijaya, Samuel and Lee, Nayeon and Dai, Wenliang and Su, Dan and Wilie, Bryan and Lovenia, Holy and Ji, Ziwei and Yu, Tiezheng and Chung, Willy and others},
  journal={arXiv preprint arXiv:2302.04023},
  year={2023}
}

@article{zheng2023judging,
  title={Judging LLM-as-a-judge with MT-bench and Chatbot Arena},
  author={Zheng, Lianmin and Chiang, Wei-Lin and Sheng, Ying and Zhuang, Siyuan and Wu, Zhanghao and Zhuang, Yonghao and Lin, Zi and Li, Zhuohan and Li, Dacheng and Xing, Eric P and others},
  journal={arXiv preprint arXiv:2306.05685},
  year={2023}
}

@article{zhou2022least,
  title={Least-to-most prompting enables complex reasoning in large language models},
  author={Zhou, Denny and Sch{\"a}rli, Nathanael and Hou, Le and Wei, Jason and Scales, Nathan and Wang, Xuezhi and Schuurmans, Dale and Cui, Claire and Bousquet, Olivier and Le, Quoc and others},
  journal={arXiv preprint arXiv:2205.10625},
  year={2022}
}

@article{madaan2023self,
  title={Self-refine: Iterative refinement with self-feedback},
  author={Madaan, Aman and Tandon, Niket and Gupta, Prakhar and Hallinan, Skyler and Gao, Luyu and Wiegreffe, Sarah and Alon, Uri and Dziri, Nouha and Prabhumoye, Shrimai and Yang, Yiming and others},
  journal={arXiv preprint arXiv:2303.17651},
  year={2023}
}

\end{document}